# Defining Relative Likelihood in Partially-Ordered Preferential Structures

Joseph Y. Halpern
IBM Research Division
Almaden Research Center, Dept. K53-B2
650 Harry Road
San Jose, CA 95120-6099
halpern@almaden.ibm.com

## Abstract

Starting with a likelihood or preference order on worlds, we extend it to a likelihood ordering on sets of worlds in a natural way, and examine the resulting logic. Lewis [1973] earlier considered such a notion of relative likelihood in the context of studying counterfactuals, but he assumed a total preference order on worlds. Complications arise when examining partial orders that are not present for total orders. There are subtleties involving the exact approach to lifting the order on worlds to an order on sets of worlds. In addition, the axiomatization of the logic of relative likelihood in the case of partial orders gives insight into the connection between relative likelihood and default reasoning.

## 1 INTRODUCTION

*Preferential structures* consist of a set of worlds partially (pre)ordered by a reflexive, transitive relation $\succeq$.[1] Various readings have been given to the $\succeq$ relation in the literature; $u \succeq v$ has been interpreted as "$u$ at least as preferred or desirable as $v$" [Kraus, Lehmann, and Magidor 1990; Doyle, Shoham, and Wellman 1991] (it is this reading that leads to the term "preferential structure"), "$u$ at least as normal (or typical) as $v$" [Boutilier 1994], and "$u$ is no more remote from actuality than $v$" [Lewis 1973]. In this paper, we focus on one other interpretation, essentially also considered by Lewis [1973]. We interpret $u \succeq v$ as meaning "$u$ is at least as likely as $v$".[2]

In the literature, preferential structures have been mainly used to give semantics to conditional logics [Lewis 1973] and, more recently, to nonmonotonic logic [Kraus, Lehmann, and Magidor 1990]. The basic modal operator in these papers has been a conditional $\to$, where $p \to q$ is interpreted as "in the most preferred/normal/likely worlds satisfying $p$, $q$ is the case". However, if we view $\succeq$ as representing likelihood, then it seems natural to define a binary operator $\gg$ on formulas such that $\varphi \gg \psi$ is interpreted as "$\varphi$ is more likely than $\psi$". Lewis [1973] in fact did define such an operator, and showed how it related to $\to$. However, he started with a *total* preorder on worlds, not a partial preorder.

In many cases in preferential or likelihood reasoning, it seems more appropriate to start with a partial preorder rather than a total preorder. As we show in this paper, there are some subtleties involved in starting with a partial preorder. What we are ultimately interested in is not an ordering on worlds, but an ordering on *sets* of worlds. To make sense of a statement like $\varphi \gg \psi$, which we read as "$\varphi$ is more likely than $\psi$", we need to compare the relative likelihood of the set of worlds satisfying $\varphi$ to that of the set satisfying $\psi$. Notice that for technical reasons (that should shortly become clear), we take as our basic primitive a strict comparison ("more likely"), rather than the nonstrict version ("at least as likely"). Given a preorder $\succeq$ on worlds, it is easy to define a strict order $\succ$ on worlds: $u \succ v$ if $u \succeq v$ and not($v \succeq u$). Moreover, it is straightforward to extend $\succeq$ and $\succ$ to preorders $\succeq^*$ and $\succ^*$ on sets of worlds. Roughly speaking, we say that $U \succeq^* V$ if for every world $v \in V$, there is a world $u \in U$ such that $u \succeq v$; we can similarly define $\succ^*$, using $\succ$. Define $U \succ' V$ if $U \succeq^* V$ and not($V \succeq^* U$). An obvious question is now whether $\succ'$ and $\succ^*$ are equivalent. It is not hard to show that they are if $\succeq$, the original preorder on worlds, is a total preorder. (This is precisely the case considered by Lewis.) On the other hand, it ing less far from actuality. Since there seems to be a split in the reading in the literature, and $\succeq$ has traditionally been taken to mean at least as likely in the literature on qualitative probability [Fine 1973; Gärdenfors 1975], we take the more natural reading here.

---

[1] A partial order $R$ is typically assumed to be reflexive, transitive, and *anti-symmetric* (so that if $R(a,b)$ and $R(b,a)$, then $a = b$). We are not assuming that $\succeq$ is anti-symmetric here, which is why it is a *preorder*.

[2] There is a tradition, starting with Lewis [1973], of taking $u \preceq v$, rather than $u \succeq v$, to mean that $u$ is as preferred or as desirable as $v$. This last reading historically comes from the interpretation of the preferred world as be-

is also not hard to construct a counterexample showing that, in general, they are different. While $U \succ^* V$ implies $U \succ' V$, the converse does not always hold. While both $\succ^*$ and $\succ'$ are reasonable notions, using $\succ^*$ allows us to make more interesting connections to conditional logic, so that is what we focus on in this paper.

More interesting observations arise when we try to axiomatize the likelihood operator. Lewis provided an axiomatization for the case of total preorders; we provide one here for the case of partial preorders. The key axioms used by Lewis were transitivity:

$$((\varphi_1 \gg \varphi_2) \wedge (\varphi_2 \gg \varphi_3)) \Rightarrow (\varphi_1 \gg \varphi_3),$$

and the *union property*:

$$(\varphi_1 \gg \varphi_2) \wedge (\varphi_1 \gg \varphi_3) \Rightarrow (\varphi_1 \gg (\varphi_2 \vee \varphi_3)).$$

This latter property is characteristic of possibility logic [Dubois and Prade 1990]. In the partially ordered case, these axioms do not suffice. We need the following axiom:

$$((\varphi_1 \vee \varphi_2 \gg \varphi_3) \wedge (\varphi_1 \vee \varphi_3 \gg \varphi_2)) \Rightarrow (\varphi_1 \gg \varphi_2 \vee \varphi_3).$$

It is not hard to show that this axiom implies transitivity and the union property (in the presence of the other axioms), but it is not equivalent to them. Interestingly, it is the property captured by this axiom that was isolated in [Friedman and Halpern 1995] as being the key feature needed for a likelihood ordering on sets to be appropriate for doing default reasoning in the spirit of [Kraus, Lehmann, and Magidor 1990]. Thus, by allowing the generality of partial preorders, we are able to clarify the connections between $\gg$, $\to$, and default reasoning.

The rest of this paper is organized as follows. In Section 2, we consider how to go from a partial preorder on worlds to a partial preorder on sets of worlds, focusing on the differences between partial and total preorders. In Section 3, we present a logic for reasoning about relative likelihood, and provide a natural complete axiomatization for it. In Section 4, we relate our results to other work on relative likelihood, as well as to work on conditional logic and nonmonotonic reasoning. We conclude in Section 5.

## 2 FROM PREORDERS ON WORLDS TO PREORDERS ON SETS OF WORLDS

We capture the likelihood ordering on a set $W$ of possible worlds by a *partial preorder*—that is, a reflexive and transitive relation—$\succeq$ on $W$. We typically write $w' \succeq w$ rather than $(w, w') \in \succeq$. As usual, we take $u \preceq v$ to be an abbreviation for $v \succeq u$, $u \succ v$ to be an abbreviation for $u \succeq v$ and not $(v \succeq u)$, and $u \prec v$ to be an abbreviation for $v \succ u$. The relation $\succ$ is a *strict partial order*, that is, it is an irreflexive (for all $w$, it is not the case that $w \succ w$) and transitive relation on $W$. We say that $\succ$ is the strict partial order *determined by* $\succeq$.

As we said in the introduction, we think of $\succeq$ as providing a likelihood, or preferential, ordering on the worlds in $W$. Thus, $w \succeq w'$ holds if $w$ is at least as likely/preferred/normal/close to actuality as $w'$. Given this interpretation, the fact that $\succeq$ is assumed to be a partial preorder is easy to justify. For example, transitivity just says that if $u$ is at least as likely as $v$, and $v$ is at least as likely as $w$, then $u$ at least as likely as $w$. Notice that since $\succeq$ is a *partial* preorder, there may be some pairs of worlds $w$ and $w'$ that are incomparable according to $\succeq$. Intuitively, we may not be prepared to say that either one is likelier than the other. We say that $\succeq$ is a *total preorder* (or *connected*, or a *linear preorder*) if for all worlds $w$ and $w'$, either $w \succeq w'$ or $w' \succeq w$.

Since we have added likelihood to the worlds, it seems reasonable to also add likelihood to the language, to allow us to say "$\varphi$ is more likely than $\psi$", for example. But what exactly should this mean? Although having $\succeq$ in our semantic model allows us to say that one world is more likely than another, it does not immediately tell us how to say that a set of worlds is more likely than another set. But, as we observed in the introduction, this is just what we need to make sense of "$\varphi$ is more likely than $\psi$".

There are a number of reasonable ways of extending the likelihood ordering on worlds to a likelihood ordering on sets. We explore one general approach here, essentially due to Lewis [1973], which has the advantage of being reasonably natural and of having some nice technical properties. Roughly speaking, we take $U$ to be more likely than $V$ if for every world in $V$, there is a more likely world in $U$.

To make this precise, first suppose that $W$ is finite. If $U, V \subseteq W$, we write $U \succeq^* V$ if for every world $v \in V$, there is a world $u \in U$ such that $u \succeq v$. It is easy to check that $\succeq^*$ as defined on finite sets is a partial preorder, that is, it is reflexive and transitive. Moreover, if $\succeq$ is a total preorder, then so is $\succeq^*$. Finally, as we would expect, we have $u \succeq v$ iff $\{u\} \succeq^* \{v\}$, so the $\succeq^*$ relation on sets of worlds can be viewed as a generalization of the $\succeq$ relation on worlds.

In a similar spirit, we can define $\succ^*$ on finite sets by taking $U \succ^* V$ to hold if $U$ is nonempty, and for every world $v \in V$, there is a world $u \in U$ such that $u \succ v$. As in the introduction, define $U \succ' V$ as an abbreviation for $U \succeq^* V$ and not $(V \succeq^* U)$. It is easy to see that $u \succ v$ iff $\{u\} \succ^* \{v\}$ iff $\{u\} \succ' \{v\}$. Thus, $\succ^*$ and $\succ'$ agree on singleton sets and extend the $\succ$ relation on worlds. Moreover, both $\succ^*$ and $\succ'$ are strict partial orders on finite sets. (The requirement that $U$ must be nonempty in the definition of $U \succ^* V$ is there to ensure that we do not have $\emptyset \succ^* \emptyset$.) As shown in Lemma 2.9, $\succ'$ and $\succ^*$ are in fact identical if the underlying preorder $\succeq$ on worlds is a total preorder.

However, as the following example shows, $\succ^*$ and $\succ'$ are not identical in general.

**Example 2.1:** Suppose $W = \{w_1, w_2\}$, and $\succeq$ is such that $w_1$ and $w_2$ are incomparable. Then it is easy to see that $\{w_1, w_2\} \succ' \{w_1\}$. However, it is not the case that $\{w_1, w_2\} \succ^* \{w_1\}$, since there is no element of $\{w_1, w_2\}$ that is strictly more likely than $w_1$. ∎

Notice that we were careful to define $\succ$ as we did only on finite sets. The following example illustrates why:

**Example 2.2:** Let $V_\infty = \{w_0, w_1, w_2, \ldots\}$, and suppose that $\succeq$ is such that
$$w_0 \prec w_1 \prec w_2 \prec \ldots$$
Then it is easy to see that if we were to apply the definition of $\succ^*$ to infinite sets, then we would have $W_\infty \succ^* W_\infty$, and $\succ$ would not be irreflexive. ∎

The approach for extending the definition of $\succ^*$ to infinite sets is also due to Lewis [1973]. The idea is to say that in order to have $U \succ^* V$, it is not enough that for every element $v$ in $V$ there is some element $u$ in $U$ that is more likely than $v$. This definition is what allows $W_\infty \succ W_\infty$ in Example 2.2. Notice that in the finite case, it is easy to see that if $U \succ^* V$, then for every element $v$ in $V$, there must some $u \in U$ such that, not only do we have $u \succ v$, but $u$ *dominates* $V$ in that, for no $v' \in V$ do we have $v' \succ u$. It is precisely this domination condition that does not hold in Example 2.2. This observation provides the motivation for the official definition of $\succeq^*$ and $\succ^*$, which applies in both finite and infinite domains.

**Definition 2.3:** Suppose $\succeq$ is a partial preorder on $W$, $U, V \subseteq W$, and $w \in W$. We say that $w$ *dominates* $V$ if for no $v \in V$ is it the case that $v \succ w$. (Notice that if $\succeq$ is a total preorder, this is equivalent to saying that $w \succeq v$ for all $v \in V$.) We write $U \succeq^* V$ if, for all $v \in V$, there exists $u \in U$ such that $u \succeq v$ and $u$ dominates $V$. We write $U \succ^* V$ if $U$ is nonempty and, for all $v \in V$, there exists $u \in U$ such that $u \succ v$ and $u$ dominates $V$. Finally, we define $U \succ' V$ if $U \succeq^* V$ and not$(V \succeq^* U)$. ∎

It is easy to see that these definitions of $\succeq^*$, $\succ^*$, and $\succ'$ agree with our earlier definitions if $U$ and $V$ are finite. We now collect some properties of $\succ^*$, $\succ'$, and $\succeq^*$. To do this, we need a few definitions.

We say that a relation $R$ on $2^W$ (not necessarily a preorder) is *qualitative* if $(V_1 \cup V_2) R V_3$ and $(V_1 \cup V_3) R V_2$ implies $V_1 R (V_2 \cup V_3)$. We say that $R$ *satisfies the union property* if $V_1 R V_2$ and $V_1 R V_3$ implies $V_1 R (V_2 \cup V_3)$. We say that $R$ is *orderly* if $U R V$, $U' \supseteq U$, and $V' \subseteq V$ implies $U' R V'$. As we now show, orderly qualitative relations always have the union property.

**Lemma 2.4:** *If $R$ is an orderly qualitative relation on $2^W$, then $R$ is transitive and satisfies the union property.*

**Proof:** Suppose $R$ is an orderly qualitative relation. To see that $R$ is transitive, suppose $V_1 R V_2$ and $V_2 R V_3$. Since $R$ is orderly, it follows that $(V_1 \cup V_3) R V_2$ and $(V_1 \cup V_2) R V_3'$. Since $R$ is qualitative, it follows that $V_1 R (V_2 \cup V_3)$. From the fact that $R$ is orderly, we get that $V_1 R V_3$. Thus, $R$ is transitive, as desired.

To see that $R$ satisfies the union property, suppose $V_1 R V_2$ and $V_1 R V_3$. Since $R$ is orderly, we have that $(V_1 \cup V_3) R V_2$ and $(V_1 \cup V_2) R V_3$. Using the fact that $R$ is qualitative, we get that $V_1 R (V_2 \cup V_3)$. Hence, $R$ satisfies the union property. ∎

The converse to Lemma 2.4 does not hold. Indeed, an orderly strict partial order on $2^W$ may satisfy the union property and still not be qualitative. For example, suppose $W = \{a, b, c\}$, and we have $\{a, b\} R \{c\}$, $\{a, c\} R \{b\}$, $\{a, b, c\} R \{b\}$, $\{a, b, c\} R \{c\}$, and $\{a, b, c\} R \{b, c\}$. It can easily be checked that $R$ is an orderly strict partial order that satisfies the union property, but is not qualitative.

With these definitions in hand, we can state the key properties of the relations we are interested in here.

**Proposition 2.5:**

*(a) If $\succeq$ is a partial preorder on $W$, then $\succeq^*$ is an orderly partial preorder on $2^W$ that satisfies the union property.*

*(b) If $\succeq$ is a partial preorder on $W$, then $\succ'$ is an orderly strict partial order on $2^W$.*

*(c) If $\succ$ is a strict partial order on $W$, then $\succ^*$ is an orderly qualitative strict partial order on $2^W$.*

**Proof:** We prove part (c) here; the proof of parts (a) and (b) is similar in spirit, and is left to the reader. The fact that $\succ^*$ is an orderly strict partial order is straightforward, and is also left to the reader. To see that $\succ^*$ is qualitative, suppose $V_1 \cup V_2 \succ^* V_3$ and $V_1 \cup V_3 \succ^* V_2$. Let $v \in V_2 \cup V_3$. We must show that there is some $v' \in V_1$ that dominates $V_2 \cup V_3$ such that $v' \succ v$. Suppose without loss of generality that $v \in V_2$ (an identical argument works if $v \in V_3$). Since $V_1 \cup V_3 \succ^* V_2$, there is some $u \in V_1 \cup V_3$ that dominates $V_2$ such that $u \succ v$. If $u$ dominates $V_3$, then it clearly dominates $V_2 \cup V_3$ and it must be in $V_1$, so we are done. Thus, we can assume that $u$ does not dominate $V_3$, so there is some element $u' \in V_3$ such that $u' \succeq u$. Since $V_1 \cup V_2 \succ^* V_3$, there must be some $v' \in V_1 \cup V_2$ such that $v'$ dominates $V_3$ and $v' \succ u'$. Since $u$ dominates $V_2$ and $u' \succeq u$, it follows that $u$ dominates $V_2$. Since $v' \succ u'$, we must have that $v'$ dominates $V_2$. Hence, $v'$ dominates $V_2 \cup V_3$. It follows that $v'$ cannot be in $V_2$, so it must be in $V_1$. Thus, we have an element in $V_1$, namely $v'$, such that $v' \succ v$ and $v'$ dominates $V_2 \cup V_3$, as desired. ∎

There are several observations worth making regard-ing this result. Clearly the union property generalizes to arbitrary finite unions. That is, if $u \succeq v_j$ for $j = 1, \ldots, N$, then $\{u\} \succeq^* \{v_1, \ldots, v_N\}$, no matter how large $N$ is, and similarly if we replace $\succeq$ by $\succ$ (since the fact that $\succ^*$ is qualitative means that it satisfies the union property, by Lemma 2.4). This is very different from probability, where sufficiently many "small" probabilities eventually can dominate a "large" probability. This suggests that $u \succ v$ should perhaps be interpreted as "$u$ is *much* more likely than $v$". In this sense, the notion of likelihood being investigated here is closer to possibility [Dubois and Prade 1990] (which also satisfies this property) than probability. Also notice that, in general, $\succ'$ does not satisfy the union property (and hence is not qualitative), and $\succeq$ is not qualitative. In Example 2.1, we have $\{w_1, w_2\} \succ' \{w_1\}$ and $\{w_1, w_2\} \succ' \{w_2\}$, but we do not have $\{w_1, w_2\} \succ' \{w_1, w_2\}$. This example also shows that $\succeq^*$ is not qualitative, since if it were, we could conclude from $\{w_1, w_2\} \succeq^* \{w_2\}$ (taking $V_1 = \{w_1\}$ and $V_2 = V_3 = \{w_2\}$ in the definition of qualitative) that $\{w_1\} \succeq^* \{w_2\}$, a contradiction.

If $\succeq$ is a total preorder, then we get further connections between these notions. Before we discuss the details, we need to define the analogue of total preorders in the strict case. A relation $R$ on an arbitrary set $W'$ (not necessarily of the form $2^W$) is *modular* if $w_1 R w_2$ implies that, for all $w_3$, either $w_3 R w_2$ or $w_1 R w_3$. Modularity is the "footprint" of a total preorder on the strict order derived from it. This is made precise in the following lemma.

**Lemma 2.6:** *If $\succeq$ is a total preorder, then the strict partial order $\succ$ determined by $\succeq$ is modular. Moreover, if $R$ is a modular, strict partial order on $W$, then there is a total preorder $\succeq$ on $W$ such that $R$ is the strict partial order determined by $\succeq$.*

**Proof:** Suppose $\succeq$ is a total preorder. To see that $\succ$ is modular, suppose that $w_1 \succ w_2$. Given an arbitrary $w_3$, if $w_3 \succeq w_1$, it follows from the transitivity of $\succeq$ that $w_3 \succ w_2$. On the other hand, if it is not the case that $w_3 \succeq w_1$, then $w_1 \succ w_3$. Thus, we have that either $w_3 \succ w_2$ or $w_1 \succ w_3$, so $\succ$ is modular.

Now suppose that $R$ is a modular strict partial order on $W$. Define $\succeq$ so that $w \succeq v$ either if $w R v$ or if neither $w R v$ nor $v R v$ hold. Clearly, $\succeq$ is reflexive. To see that it is transitive, suppose that $v_1 \succeq v_2$ and $v_2 \succeq v_3$. There are three cases: (1) If $v_1 R v_2$, then since $R$ is modular, we have that either $v_1 R v_3$ or $v_3 R v_2$. We cannot have $v_3 R v_2$, for then we would not have $v_2 \succeq v_3$. Thus, we must have $v_1 R v_3$, and hence $v_1 \succeq v_3$. (2) If $v_2 R v_3$, then using modularity again, we get that either $v_1 R v_3$ or $v_2 R v_1$. Again, we cannot have $v_2 R v_1$, so we must have $v_1 R v_3$, and so we also have $v_1 \succeq v_3$. (3) If neither $v_1 R v_2$ nor $v_2 R v_3$ hold, then we claim that neither $v_1 R v_3$ nor $v_3 R v_1$ hold. For if $v_1 R v_3$, then by modularity, we must have either $v_1 R v_2$ or $v_2 R v_3$. And if $v_3 R v_1$, then either $v_3 R v_2$ or $v_2 R v_1$, which contradicts the assumption that $v_1 \succeq v_2$ and $v_2 \succeq v_3$. Thus, we can again conclude that $v_1 \succeq v_3$. Thus, $\succeq$ is transitive. Finally, it is almost immediate from the definition that $R$ is the strict partial order determined by $\succeq$. ∎

Modularity is preserved when we lift the preorder from $W$ to $2^W$.

**Lemma 2.7:** *If $\succ$ is a modular relation on $W$, then $\succ^*$ is a modular relation on $2^W$.*

**Proof:** Suppose $\succ$ is modular. We want to show that $\succ^*$ is modular. So suppose that $V_1 \succ^* V_2$, and it is not the case that $V_1 \succ^* V_3$. We must show that $V_3 \succ^* V_2$. Since it is not the case that $V_1 \succ^* V_3$, there must be some $v^* \in V_3$ such that for all $u \in V_1$, we do not have $u \succ v^*$. Now suppose $v \in V_2$. We claim that $v^* \succ v$. To see this, note that since $V_1 \succ^* V_2$, there must be some $u^* \in V_1$ such that $u^* \succ v$. Since $\succ$ is modular, we have that either $u^* \succ v^*$ or $v^* \succ v$. Since, by choice of $v^*$, we do not have $u^* \succ v^*$, we must have $v^* \succ v$. It follows that $V_3 \succ^* V_2$. ∎

Although we showed that the converse to Lemma 2.4 does not hold in general for strict partial orders, it does hold for orders that are modular.

**Lemma 2.8:** *If $R$ is a modular strict partial order and satisfies the union property, then $R$ is qualitative.*

**Proof:** Suppose that $R$ is modular strict partial order that satisfies the union property. To see that $R$ is qualitative, suppose that $(V_1 \cup V_2) R V_3$ and $(V_1 \cup V_3) R V_2$. Since $R$ is modular, it follows that either $(V_1 \cup V_2) R V_1$ or $V_1 R V_3$. If $(V_1 \cup V_2) R V_1$, then, using the fact that $R$ satisfies the union property and $(V_1 \cup V_2) R V_3$, we get that $(V_1 \cup V_2) R (V_1 \cup V_3)$. Using transitivity, it follows that $(V_1 \cup V_2) R V_2$. Using the union property again, we get that $(V_1 \cup V_2) R (V_1 \cup V_2)$. This contradicts the assumption that $R$ is irreflexive. Thus, we must have that $V_1 R V_3$. A similar argument shows that $V_1 R V_2$. Using the union property, we get that $V_1 R (V_2 \cup V_3)$, as desired. ∎

As shown in [Friedman and Halpern 1995], there is a connection between nonmonotonic reasoning, conditional logic, and the qualitative property. (This is discussed in Section 4.) It is because of this relationship that we consider $\succ^*$ rather than $\succeq^*$ or $\succ'$. Lewis [1973] was able to use $\succ'$ because he focused on total preorders. The following lemma makes this precise.

**Lemma 2.9:** *If $\succeq$ is a total preorder, then $\succ^*$ and $\succ'$ agree. In general, $U \succ^* V$ implies $U \succ' V$, but the converse does not hold.*

**Proof:** It is immediate from the definitions that $U \succ^* V$ implies $U \succ' V$, and the fact that the converse does not hold is shown by Example 2.1. To show that $\succ^*$ and $\succ'$ are equivalent if $\succeq$ is a total preorder, suppose $U \succ' V$. Clearly $U$ is nonempty, since $V \succeq^* \emptyset$ for all $V$. We want to show that $U \succ^* V$, so we must show that for all $v \in V$, there is some $u \in U$ that dominates $V$ such that $u \succ v$. Given $v \in V$, since $U \succeq^* V$, there must be some $u \in U$ that dominates $V$ such that $u \succeq v$. If $u \succ v$, then we are done. If not, then $v \succeq u$. Since it is not the case that $V \succeq U$, there must be some $u' \in U$ such that it is not the case that $v \succeq u'$. Since $\succeq$ is a total order, we must have $u' \succ v$. Since $v \succeq u$, we also have $u' \succ u$. Since $u$ dominates $V$, so does $u'$. It follows that $U \succ^* V$, as desired. ∎

We close this section by considering when a preorder on $2^W$ can be viewed as being generated by a preorder on $W$. This result turns out to play a key role in our completeness proof, and emphasizes the role of the qualitative property.

**Theorem 2.10:** *Let $\mathcal{F}$ be a finite algebra of subsets of $W$ (that is, $\mathcal{F}$ is a set of subsets of $W$ that is closed under union and complementation and contains $W$ itself) and $R$ an orderly qualitative relation on $\mathcal{F}$.*

(a) *If $R$ is a total preorder on $\mathcal{F}$, then there is a total preorder $\succeq$ on $W$ such that $R$ and $\succeq^*$ agree on $\mathcal{F}$ (that is, for $U, V \in \mathcal{F}$, we have $U R V$ iff $U \succeq^* V$).*

(b) *If $R$ is a strict partial order and each nonempty set in $\mathcal{F}$ has at least $2^{|\mathcal{F}|^{\log \log(|\mathcal{F}|)}}$ elements, then there is a partial preorder $\succ$ on $W$ such that $R$ and $\succ^*$ agree on $\mathcal{F}$.*

**Proof:** An *atom* of $\mathcal{F}$ is a minimal nonempty element of $\mathcal{F}$. Since $\mathcal{F}$ is finite, it is easy to see that every element of $\mathcal{F}$ can be written as a union of atoms, and the atoms are disjoint. Part (a) is easy: for each $w \in W$, let $A_w$ be the unique atom in $\mathcal{F}$ containing $w$. Define $\succeq$ on $W$ so that $v \succeq w$ iff $A_v R A_w$. It is easy to see that if $R$ is a total preorder on $\mathcal{F}$, then $\succeq$ is a total preorder on $W$ and $R$ agrees with $\succeq^*$ on $\mathcal{F}$. The proof of (b) is considerably more difficult; we leave details to the full paper. ∎

It is not clear that the requirement that the sets in $\mathcal{F}$ have at least $2^{|\mathcal{F}|^{\log \log(|\mathcal{F}|)}}$ elements is necessary. However, it can be shown that Theorem 2.10(b) does not hold without some assumption on the cardinality of elements in $\mathcal{F}$. For example, suppose that the atoms of $\mathcal{F}$ are $A$, $B$, and $C$. Let $R$ be defined so that the only sets related by $R$ are $(B \cup C) R A$, $W R A$, and $X R \emptyset$ for all nonempty $X \in \mathcal{F}$. It is easy to see that $R$ is a strict partial order, $R$ is orderly, and $R$ is qualitative. However, if $W = \{a, b, c\}$, $A = \{a\}$, $B = \{b\}$, and $C = \{c\}$, there is no ordering $\succ$ on $W$ such that $\succ^*$ and $R$ agree on $\mathcal{F}$: it is easy to see that such an ordering $\succ$ must make $a$, $b$, and $c$ incomparable. But if they are incomparable, we cannot have $\{b, c\} \succ^* \{a\}$. On the other hand, if we allow $C$ to have two elements, by taking $W = \{a, b, c, d\}$, $A = \{a\}$, $B = \{b\}$, and $C = \{c, d\}$, then there is an ordering $\succ$ such that $\succ^* = R$: we simply take $a \succ c$ and $b \succ d$.

## 3 A LOGIC OF RELATIVE LIKELIHOOD

We now consider a logic for reasoning about relative likelihood. Let $\Phi$ be a set of primitive propositions. A *basic likelihood formula (over $\Phi$)* is one of the form $\varphi \gg \psi$, where $\varphi$ and $\psi$ are propositional formulas over $\Phi$. We read $\varphi \gg \psi$ as "$\varphi$ is more likely than $\psi$". Let $\mathcal{L}$ consist of Boolean combinations of basic likelihood formulas. Notice that we do not allow nesting of likelihood in $\mathcal{L}$, nor do we allow purely propositional formulas. There would no difficulty extending the syntax and semantics to deal with them, but this would just obscure the issues of interest here.

A *preferential structure* (over $\Phi$) is a tuple $M = (W, \succeq, \pi)$, where $W$ is a (possibly infinite) set of possible worlds, $\succeq$ is a partial preorder on $W$, and $\pi$ associates with each world in $W$ a truth assignment to the primitive propositions in $\Phi$. Notice that there may be two or more worlds with the same truth assignment. As we shall see, in general, we need to have this, although in the case of total preorders, we can assume without loss of generality that there is at most one world associated with each truth assignment.

We can give semantics to formulas in $\mathcal{L}$ in preferential structures in a straightforward way. For a propositional formula $\varphi$, let $[\![\varphi]\!]_M$ consist of the worlds in $M$ whose truth assignment satisfies $\varphi$. We then define

$$M \models \varphi \gg \psi \text{ if } [\![\varphi]\!]_M \succ^* [\![\psi]\!]_M.$$

We extend $\models$ to Boolean combinations of basic formulas in the obvious way.

Notice that $M \models \neg(\neg\varphi \gg \mathit{false})$ iff $[\![\neg\varphi]\!]_M = \emptyset$ iff $[\![\varphi]\!]_M = W$. Let $K\varphi$ be an abbreviation for $\neg(\neg\varphi \gg \mathit{false})$. It follows that $M \models K\varphi$ iff $\varphi$ is true at all possible worlds.[3]

With these definitions, we can provide a sound and complete axiomatization for this logic of relative likelihood. Let AX consist of the following axioms and inference rules.

L1. All substitution instances of tautologies of propositional calculus

L2. $\neg(\varphi \gg \varphi)$

L3. $((\varphi_1 \vee \varphi_2 \gg \varphi_3) \wedge (\varphi_1 \vee \varphi_3 \gg \varphi_2)) \Rightarrow (\varphi_1 \gg \varphi_2 \vee \varphi_3)$

L4. $(K(\varphi \Rightarrow \varphi') \wedge K(\psi' \Rightarrow \psi) \wedge (\varphi \gg \psi)) \Rightarrow \varphi' \gg \psi'$

MP. From $\varphi$ and $\varphi \Rightarrow \psi$ infer $\psi$ (Modus ponens)

Gen. From $\varphi$ infer $K_i\varphi$ (Knowledge Generalization)

Note that L2, L3, and L4 just express the fact that $\succ^*$ is irreflexive, orderly, and qualitative, respectively.

---
[3] $K$ was defined by Lewis [1973], although he used □.

**Theorem 3.1:** *AX is a sound and complete axiomatization of the language $\mathcal{L}$ with respect to preferential structures.*

**Proof:** The soundness of L1 is immediate. It is clear that the fact that $\succ^*$ is irreflexive and qualitative, as shown in Proposition 2.5, implies that L2 and L3 are sound. To see that L4 corresponds to orderliness, note that if $M \models K(\varphi \Rightarrow \varphi') \wedge K(\psi' \Rightarrow \psi)$ and $\varphi \gg \psi$, then $[\![\varphi]\!]_M \subseteq [\![\varphi']\!]_M$, $[\![\psi']\!]_M \subseteq [\![\psi]\!]_M$, and $[\![\varphi]\!]_M \succ^* [\![\psi]\!]_M$. Since $\succ^*$ is orderly, it follows that $[\![\varphi']\!]_M \succ^* [\![\psi']\!]_M$, so $M \models \varphi' \gg \psi'$. Thus, L4 is sound. It is also clear that MP and Gen preserve validity.

The completeness proof starts out, as is standard for completeness proofs in modal logic, with the observation that it suffices to show that a consistent formula is satisfiable. That is, we must show that for every formula $\varphi$ for which it is not the case that $\neg\varphi$ is provable from AX is satisfiable in some preferential structure $M$. However, the standard modal logic techniques of constructing a *canonical* model (see, for example, [Hughes and Cresswell 1968]) do not seem to work in this case. Finding an appropriate partial preorder is nontrivial. For this we use (part (b) of) Theorem 2.10. We leave details to the full paper. ∎

What happens if we start with a total preorder? Let $AX^M$ consist of AX together with the obvious axiom expressing modularity:

L5. $\varphi_1 \gg \varphi_2 \Rightarrow ((\varphi_1 \gg \varphi_3) \vee (\varphi_3 \gg \varphi_2))$

We say that a preferential structure is *totally preordered* if it has the form $(W, \succeq, \pi)$, where $\succeq$ is a total preorder on $W$.

**Theorem 3.2:** *$AX^M$ is a sound and complete axiomatization of the language $\mathcal{L}$ with respect to totally preordered preferential structures.*

We remark that in light of Proposition 2.4, we can replace L4 in $AX^M$ by axioms saying that $\gg$ is transitive and satisfies the union property, namely:

L6. $((\varphi_1 \gg \varphi_2) \wedge (\varphi_2 \gg \varphi_3)) \Rightarrow (\varphi_1 \gg \varphi_3)$
L7. $((\varphi_1 \gg \varphi_2) \vee (\varphi_1 \gg \varphi_3)) \Rightarrow (\varphi_1 \gg \varphi_1 \vee \varphi_2)$

The result is an axiomatization that is very similar to that given by Lewis [1973].

In the proof of Theorem 3.1, when showing that a consistent formula $\varphi$ is satisfiable, the structure constructed may have more than one world with the same truth assignment. This is necessary, as the following example shows. (We remark that this observation is closely related to the cardinality requirements in Theorem 2.10(b).)

**Example 3.3:** Suppose $\Phi = \{p, q\}$. Let $\varphi$ be the formula $(p \gg \neg p \wedge q) \wedge \neg (p \wedge q \gg \neg p \wedge q) \wedge \neg (p \wedge \neg q \gg \neg p \wedge q)$. It is easy to see that $\varphi$ is satisfied in a structure consisting of four worlds, $w_1, w_2, w_3, w_4$, such that $w_1 \succ w_3$, $w_2 \succ w_4$, $p \wedge q$ is true at $w_1$, $p \wedge \neg q$ is true at $w_2$, and $\neg p \wedge q$ is true at both $w_3$ and $w_4$. However, $\varphi$ is not satisfiable in any structure where there is at most one world satisfying $\neg p \wedge q$. For suppose $M$ were such a structure, and let $w$ be the world in $M$ satisfying $\neg p \wedge q$. Since $M \models p \gg \neg p \wedge q$, it must be the case that $[\![p]\!]_M \succ^* \{w\}$. Thus, there must be a world $w' \in [\![p]\!]_M$ such that $w' \succ w$. But $w'$ must satisfy one of $p \wedge q$ or $p \wedge \neg q$, so $M \models (p \wedge q \gg \neg p \wedge q) \vee (p \wedge \neg q \gg \neg p \wedge q)$, contradicting the assumption that $M \models \varphi$. ∎

It is not hard to see that the formula $\varphi$ of Example 3.3 is not satisfiable in a totally preordered preferential structure. This is not an accident.

**Proposition 3.4:** *If a formula is satisfiable in a totally preordered preferential structure, then it is satisfiable in a totally preordered preferential structure with at most one world per truth assignment.*

The results of this and the previous section help emphasize the differences between totally preordered and partially preordered structures.

## 4  RELATED WORK

The related literature basically divides into two groups (with connections between them): other approaches to relative likelihood, and work on conditional and nonmonotonic logic.

We first consider relative likelihood. Gärdenfors [1975] considered a logic of relative likelihood, but he took as primitive a total preorder on the sets in $2^W$, and focused on connections with probability. In particular, he added axioms to ensure that, given a preorder $\succeq^*$ on $2^W$, there was a probability function Pr with the property that (in our notation) $U \succeq^* V$ iff $\Pr(U) \geq \Pr(V)$. Fine [1973] defines a qualitative notion $\preceq$ of *comparative probability*, but like Gärdenfors, assumes that the preorder on sets is primitive, and is largely concerned with connections to probability.

Halpern and Rabin [1987] consider a logic of likelihood where absolute statements about likelihood can be made ($\varphi$ is likely, $\psi$ is somewhat likely, and so on), but there is no notion of relative likelihood.

Of course, there are many more quantitative notions of likelihood, such as probability, possibility [Dubois and Prade 1990], ordinal conditional functions (OCFs) [Spohn 1987], and Dempster-Shafer belief functions [Shafer 1976]. The ones closest to the relative likelihood considered here are possibility and OCFs. Recall that a possibility measure Poss on $W$ associates with each world its *possibility*, a number in $[0, 1]$, such that for $V \subseteq W$, we have $\text{Poss}(V) = \sup\{\text{Poss}(v) : v \in V\}$, with the requirement that $\text{Poss}(W) = 1$. Clearly a possibility measure places a total preorder on sets, and satisfies the union property, since $\text{Poss}(A \cup B) = \max(\text{Poss}(A), \text{Poss}(B))$. The same is true for OCFs;

we refer the reader to [Spohn 1987] for details. Fariñas del Cerro and Herzig [1991] define a logic QPL (Qualitative Possibilistic Logic) with a modal operator $\triangleleft$, where $\varphi \triangleleft \psi$ is interpreted as $\text{Poss}(\llbracket\varphi\rrbracket) \leq \text{Poss}(\llbracket\psi\rrbracket)$. Clearly, $\varphi \triangleleft \psi$ essentially corresponds to $\psi \gg \varphi$. They provide a complete axiomatization for their logic; further discussion of the logic can be found in [Bendová and Hájek 1993]. Not surprisingly, an analogue of $AX^M$ is also complete for the logic. We discuss further connections between possibility measures, OCFs, and our logic below, in the context of conditionals.

Finally, we should mention the work of Doyle, Shoham, and Wellman [1991]. They define a logic of relative desire, starting with a preference order on worlds. They extend this to a preference order on sets, but in a much different way than we do. Using $\geq$ for their notion, they define $U \geq V$ if $u \succeq v$ for every $u \in U$ and $v \in V$; a strict notion $>$ can be similarly defined. It is easy to see that $U > V$ is not equivalent to $U \geq V$ and $\text{not}(U \geq V)$, even if $\succeq$ is a total preorder, although it implies it. Clearly, $U \geq V$ implies $U \succeq^* V$ and $U > V$ implies $U \succ^* V$, but the converse does not hold. The requirements for $\geq$ and $>$ are significantly stronger than those for $\succeq^*$ and $\succ^*$. As Doyle, Shoham, and Wellman themselves point out, these relations are too weak to allow us to make important distinctions. They go on to define other notions of comparison, but these are incomparable to the notions considered here, and are more tuned to their applications.

We now turn our attention to conditional logic. Lewis's main goal in considering preferential structures was to capture a counterfactual conditional $\rightarrow$, where $\psi \rightarrow \varphi$ is read as "if $\psi$ were the case, then $\varphi$ would be true" as in "if kangaroos had no tails, then they would topple over". He takes this to be true at a world $w$ if, in all the worlds "closest" to $w$ (where closeness is defined by a preorder $\succeq$) where kangaroos don't have tails, it is the case that kangaroos topple over.[4]

More abstractly, in the case where $W$ is finite, for a subset $V \subseteq W$, let $\min(V) = \{v \in V : v' \succ v \text{ implies } v' \notin V\}$. Thus, $\min(V)$ consists of all worlds $v \in V$ such that no world $v' \in V$ is considered more likely than $v$. (We take $\min(\emptyset) = \emptyset$.)

If $W$ is finite, we define

$$(M, w) \models \psi \rightarrow \varphi \text{ if } \min(\llbracket\psi\rrbracket_M) \subseteq \llbracket\varphi\rrbracket_M.$$

Thus, $\psi \rightarrow \varphi$ is true exactly if $\varphi$ is true at the most likely (or closest) worlds where $\varphi$ is true.

For infinite domains, this definition does not quite capture our intentions. For example, in Example 2.2, we have $\min(W_\infty) = \emptyset$. It follows that if $M = (W_\infty, \succeq, \pi)$, then $M \models true \rightarrow \neg p$ even if $\pi$ makes $p$ true at every world in $W_\infty$. We certainly would not want to say that "if $true$ were the case, then $p$ would be false" is true if $p$ is true at all the worlds in $W_\infty$! The solution here, again due to Lewis, is much like that for $\succ^*$ in infinite domains. We then say $M \models \psi \rightarrow \varphi$ if for all $u \in \llbracket\psi\rrbracket_M$, there exists a world $v \in \llbracket\varphi \wedge \psi\rrbracket_M$ such that $v \succeq u$ and $v$ dominates $\llbracket\varphi \wedge \neg\psi\rrbracket_M$. This definition can be shown to agree with the definition for the case of finite $W$. Lewis [1973] argues that it captures many of our intuitions for counterfactual reasoning.

---

[4]To really deal appropriately with counterfactuals, we require not one preorder $\succeq$, but a possibly different preorder $\succeq_w$ for each world $w$, since the notion of closeness in general depends on the actual world. We ignore this issue here, since it is somewhat tangential to our concerns.

We can give $\rightarrow$ another interpretation, perhaps more natural if we are thinking in terms of likelihood. We often want to say that $\varphi$ is more likely than not—in $\mathcal{L}$, this can be expressed as $\varphi \gg \neg\varphi$. More generally, we might want to say that *relative to* $\psi$, or *conditional on* $\psi$ *being the case*, $\varphi$ is more likely than not. By this we mean that if we restrict to worlds where $\psi$ is true, $\varphi$ is more likely than not, that is, the worlds where $\varphi \wedge \psi$ is true are more likely than the worlds where $\neg\varphi \wedge \psi$ is true.

Let us define $\psi \rightarrow' \varphi$ to be an abbreviation for $K\neg\psi \vee (\varphi \wedge \psi \gg \neg\varphi \wedge \psi)$. That is, $\psi \rightarrow' \varphi$ is true vacuously in a structure $M$ if $\psi$ does not hold in any world in $M$; otherwise, it holds if $\varphi$ is more likely than not in the worlds satisfying $\psi$.

Although the intuition for $\rightarrow'$ seems, on the surface, quite different from that for $\rightarrow$, especially in finite domains, it is easy to see that they are equivalent. (This connection between $\rightarrow$ and $\rightarrow'$ was already observed by Lewis [1973] in the case of total preorders.)

**Lemma 4.1:** *For all structures* $M$, *we have* $M \models \varphi \rightarrow \psi$ *iff* $M \models \varphi \rightarrow' \psi$.

Given Lemma 4.1, we can write $\rightarrow$ for both $\rightarrow$ and $\rightarrow'$. The lemma also allows us to apply the known results for conditional logic to the logic of relative likelihood defined here. In particular, the results of [Friedman and Halpern 1994] show that the validity problem for the logic of Section 3 is co-NP complete, no harder than that of propositional logic, for the case of both partial and total preorders.[5]

More recently, $\rightarrow$ has been used to capture nonmonotonic default reasoning [Kraus, Lehmann, and Magidor 1990; Boutilier 1994]. In this case, a statement like $Bird \rightarrow Fly$ is interpreted as "birds typically fly", or "by default, birds fly". The semantics does not change: $Bird \rightarrow Fly$ is true if in the most likely worlds satisfying $Bird$, $Fly$ holds as well. Dubois and Prade [1991] have shown that possibility can be used to give semantics to defaults as well, where $\psi \rightarrow \varphi$ is interpreted as $\text{Poss}(\psi) = 0$ or $\text{Poss}(\varphi \wedge \psi) > \text{Poss}(\varphi \wedge \neg\psi)$. Of course, this is just the analogue of the definition of $\rightarrow$ in terms of $\succ^*$. Goldszmidt and Pearl [1992] have shown that a similar approach works if we use Spohn's OCFs.

---

[5]We remark that there are also well known axiomatizations for various conditional logics [Burgess 1981; Friedman and Halpern 1994; Lewis 1973]. These do not immediately give us a complete axiomatization for the logic of relative likelihood considered here, since we must find axioms in the language with $\gg$, not in the language with $\rightarrow$.

These results are clarified and unified in [Friedman and Halpern 1995]. Suppose we start with some mapping Pl of sets to a partially ordered space with minimal element $\perp$ (such a mapping is called a *plausibility measure* in [Friedman and Halpern 1995]). Define $\psi \rightarrow \varphi$ as $\text{Pl}(\varphi) = \perp$ or $\text{Pl}(\psi \wedge \varphi) > \text{Pl}(\psi \wedge \neg\varphi)$. Then it is shown that $\rightarrow$ satisfies the *KLM properties*—the properties isolated in [Kraus, Lehmann, and Magidor 1990] as forming the core of default reasoning—if and only if Pl is qualitative, at least when restricted to disjoint sets.[6] Since $\succeq^*$, Poss, and OCFs give rise to qualitative orders on $2^W$, it is no surprise that they should all lead to logics that satisfy the KLM properties.

## 5  CONCLUSION

We have investigated a notion of relative likelihood starting with a preferential ordering on worlds. This notion was earlier studied by Lewis [1973] in the case where the preferential order is a total preorder; the focus of this paper is on the case where the preferential order is a partial preorder. Our results show that there are significant differences between the totally ordered and partially ordered case. By focusing on the partially ordered case, we bring out the key role of the qualitative property (axiom L3), whose connections to conditional logic were already observed in [Friedman and Halpern 1995].

### Acknowledgments

I'd like to thank Nir Friedman for many interesting and useful discussions on plausibility that formed the basis for this paper, and for pointing out the reference [Doyle, Shoham, and Wellman 1991]. This work was supported in part by NSF grant IRI-93-03109.

## References

Bendová, K. and P. Hájek (1993). Possibilistic logic as tense logic. In N. P. Carreté and M. G. Singh (Eds.), *Qualitative Reasoning and Decision Technologies*, pp. 441–450.

Boutilier, C. (1994). Conditional logics of normality: a modal approach. *Artificial Intelligence 68*, 87–154.

Burgess, J. (1981). Quick completeness proofs for some logics of conditionals. *Notre Dame Journal of Formal Logic 22*, 76–84.

Doyle, J., Y. Shoham, and M. P. Wellman (1991). A logic of relative desire. In *Proc. 6th International Symposium on Methodologies for Intelligent Systems*, pp. 16–31.

Dubois, D. and H. Prade (1990). An introduction to possibilistic and fuzzy logics. In G. Shafer and J. Pearl (Eds.), *Readings in Uncertain Reasoning*, pp. 742–761. Morgan Kaufmann.

Dubois, D. and H. Prade (1991). Possibilistic logic, preferential models, non-monotonicity and related issues. In *Proc. Twelfth International Joint Conference on Artificial Intelligence (IJCAI '91)*, pp. 419–424.

Fariñas del Cerro, L. and A. Herzig (1991). A modal analysis of possibilistic logic. In *Symbolic and Quantitative Approces to Uncertainty*, Lecture Notes in Computer Science, Vol. 548, pp. 58–62. Springer-Verlag.

Fine, T. L. (1973). *Theories of Probability*. Academic Press.

Friedman, N. and J. Y. Halpern (1994). On the complexity of conditional logics. In *Principles of Knowledge Representation and Reasoning: Proc. Fourth International Conference (KR '94)*, pp. 202–213.

Friedman, N. and J. Y. Halpern (1995). Plausibility measures and default reasoning. In *Proc. National Conference on Artificial Intelligence (AAAI '96)*. Also available at http://robotics.stanford.edu/users/nir.

Gärdenfors, P. (1975). Qualitative probability as an intensional logic. *Journal of Philosophical Logic 4*, 171–185.

Goldszmidt, M. and J. Pearl (1992). Rank-based systems: A simple approach to belief revision, belief update and reasoning about evidence and actions. In *Principles of Knowledge Representation and Reasoning: Proc. Third International Conference (KR '92)*, pp. 661–672.

Halpern, J. Y. and M. O. Rabin (1987). A logic to reason about likelihood. *Artificial Intelligence 32*(3), 379–405.

Hughes, G. E. and M. J. Cresswell (1968). *An Introduction to Modal Logic*. Methuen.

Kraus, S., D. Lehmann, and M. Magidor (1990). Nonmonotonic reasoning, preferential models and cumulative logics. *Artificial Intelligence 44*, 167–207.

Lewis, D. K. (1973). *Counterfactuals*. Harvard University Press.

Shafer, G. (1976). *A Mathematical Theory of Evidence*. Princeton University Press.

Spohn, W. (1987). Ordinal conditional functions: a dynamic theory of epistemic states. In W. Harper and B. Skyrms (Eds.), *Causation in Decision, Belief Change and Statistics*, Volume 2, pp. 105–134. Reidel.

---

[6]That is, if $V_1$, $V_2$, and $V_3$ are *disjoint* sets, we require that if $\text{Pl}(V_1 \cup V_2) > \text{Pl}(V_3)$ and $\text{Pl}(V_1 \cup V_3) > \text{Pl}(V_2)$, then $\text{Pl}(V_1) > \text{Pl}(V_2 \cup V_3)$. The result also requires the assumption that if $\text{Pl}(U) = \text{Pl}(V) = \perp$, then $\text{Pl}(U \cup V) = \perp$.